\documentclass[letterpaper, 10 pt, conference]{ieeeconf}  

\IEEEoverridecommandlockouts                              

\title{\LARGE \bf
Autonomous Human-Robot Interaction via Operator Imitation
}

\author{Sammy Christen$^{1}$, David Müller$^{1}$, Agon Serifi$^{1}$, Ruben Grandia$^{1}$, \\ Georg Wiedebach$^{2}$, Michael A. Hopkins$^{2}$, Espen Knoop$^{1}$, Moritz Bächer$^{1}$
\thanks{$^{1}$Disney Research, Switzerland. $^{2}$Disney Research, USA}%
}

\usepackage{amsmath}
\usepackage{booktabs}
\usepackage{graphicx}
\usepackage{tabularx} 
\usepackage{colortbl}
\usepackage{pgf}
\usepackage{pgfplots}
\usepackage{multirow}  
\usepackage{siunitx}
\usepackage{hyperref}

\begin{document}

\maketitle
\thispagestyle{empty}
\pagestyle{empty}

\newcommand{\methodname}{AutoOp}

\newcommand{\figref}[1]{Fig.~\ref{#1}}
\newcommand{\secref}[1]{Sec.~\ref{#1}}
\newcommand{\algref}[1]{Algorithm~\ref{#1}}
\newcommand{\eqnref}[1]{Eq.~\eqref{#1}}
\newcommand{\tabref}[1]{Tab.~\ref{#1}}

\newcommand{\R}{\rm I\!R}

\newcommand\blfootnote[1]{%
  \begingroup
  \vspace{-4mm}
  \renewcommand\thefootnote{}\footnote{#1}%
  \addtocounter{footnote}{-1}%
  \endgroup
}

\makeatletter
\def\onedot{\ifx\@let@token.\else.\null\fi}
\makeatother
\def\etal{\emph{et al}\onedot}

\newcommand{\TODO}[1]{\textbf{\color{red}[TODO: #1]}}

\newif\ifshowcomments
\showcommentstrue 

\ifshowcomments
        \newcommand{\note}[3]{{\textcolor{#2}{[#1: #3]}}}
        \newcommand{\moritz}[1]{\note{Moritz}{green}{#1}}
	\newcommand{\sammy}[1]{\note{Sammy}{teal}{#1}}
        \newcommand{\mike}[1]{{\color{blue} [Mike: #1] }}
        \newcommand{\david}[1]{{\color{red} [David: #1] }}
        \newcommand{\agon}[1]{{\color{magenta} [Agon: #1] }}
        \newcommand{\ruben}[1]{{\color{olive} [Ruben: #1] }}
        \newcommand{\georg}[1]{{\color{cyan} [Georg: #1] }}

\else
        \newcommand{\note}[3]{\unskip}
        \newcommand{\moritz}[1]{\unskip}
        \newcommand{\sammy}[1]{\unskip}
	\newcommand{\mike}[1]{\unskip}
	\newcommand{\david}[1]{\unskip}
	\newcommand{\agon}[1]{\unskip}
	\newcommand{\ruben}[1]{\unskip}
	\newcommand{\georg}[1]{\unskip}
\fi

\newcommand{\colorcell}[1]{%
    \pgfmathsetmacro{\greencomp}{ 1 - 0.6 * #1}  
    \pgfmathsetmacro{\redblue}{ 1- 0.3 * #1}  
    \xdef\cellcolorval{\greencomp,\greencomp,\redblue}  
    \cellcolor[rgb]{\cellcolorval} #1~~~  
}

\begin{abstract}

Teleoperated robotic characters can perform expressive interactions with humans, relying on the operators' experience and social intuition. In this work, we propose to create autonomous interactive robots, by training a model to imitate operator data. Our model is trained on a dataset of human-robot interactions, where an expert operator is asked to vary the interactions and mood of the robot, while the operator commands as well as the pose of the human and robot are recorded. Our approach learns to predict continuous operator commands through a diffusion process and discrete commands through a classifier, all unified within a single transformer architecture. We evaluate the resulting model in simulation and with a user study on the real system. We show that our method enables simple autonomous human-robot interactions that are comparable to the expert-operator baseline, and that users can recognize the different robot moods as generated by our model. Finally, we demonstrate a zero-shot transfer of our model onto a different robotic platform with the same operator interface.

\end{abstract}
\section{Introduction}

Today, a wide variety of robotic systems are capable of expressing a rich set of behaviors, including quadrupeds \cite{hutter2016anymal, bostondynamics_spot}, humanoids \cite{cheng2024express}, and non-anthropomorphic robots \cite{grandia2024design}. These systems are suitable for human-robot interaction (HRI), in particular when controlled by a skilled operator.

Operators can assess the environment, interpret the behavior of the human, and initiate appropriate interactions effectively.  
A key challenge is achieving autonomous HRI without an operator in the loop.

\blfootnote{This work has been submitted to the IEEE for possible publication. Copyright may be transferred without notice, after which this version may no longer be accessible.}

Full autonomy in HRI combines decision-making, motion control, and social interactions \cite{thomaz2016hri, sheridan2016human}. Furthermore, it requires incorporating aspects of the theory of mind \cite{frith2005theory}, such as understanding intent, beliefs, emotional states, and desires—both of the robot itself and the human. While such capabilities remain an open challenge, we observe that human operators already enable robots to convey expressive and lifelike behaviors through their experience and simple social cues such as human movements. For example, a shy robot might follow the human while keeping a safe distance and looking towards the ground, or a joyful robot might run closely behind the human and initiate dancing motions from time to time. Inspired by this, we aim to develop a framework that allows robots to autonomously engage in interactions—without contact—by approximating the human state through the robot-relative human pose, and express different moods at a level comparable to a human operator.

Developing an expressive and autonomous system for HRI requires methods that are flexible, expressive, and scalable. Heuristic-based approaches \cite{bauer2009heuristic, block2022huggiebot} are inherently limited in scalability, require expert knowledge, and may need to be redefined for each robot. Similar challenges arise when defining rewards for a reinforcement learning policy \cite{qureshi2018intrinsically}, with the additional difficulty of realistically simulating human motions \cite{thumm2024human} or long training times in the real world~\cite{qureshi2016robot}. An alternative is to collect real-world data and train supervised models, a popular approach in robotic manipulation \cite{black2024pi0, chi2023diffusion}. In the context of HRI, collecting such data can be time-consuming. Recent advances in user-controllable robots via gamepads offer a promising approach to simplifying data collection in such settings.

\begin{figure}[t]
    \centering
    \includegraphics[width=\linewidth]{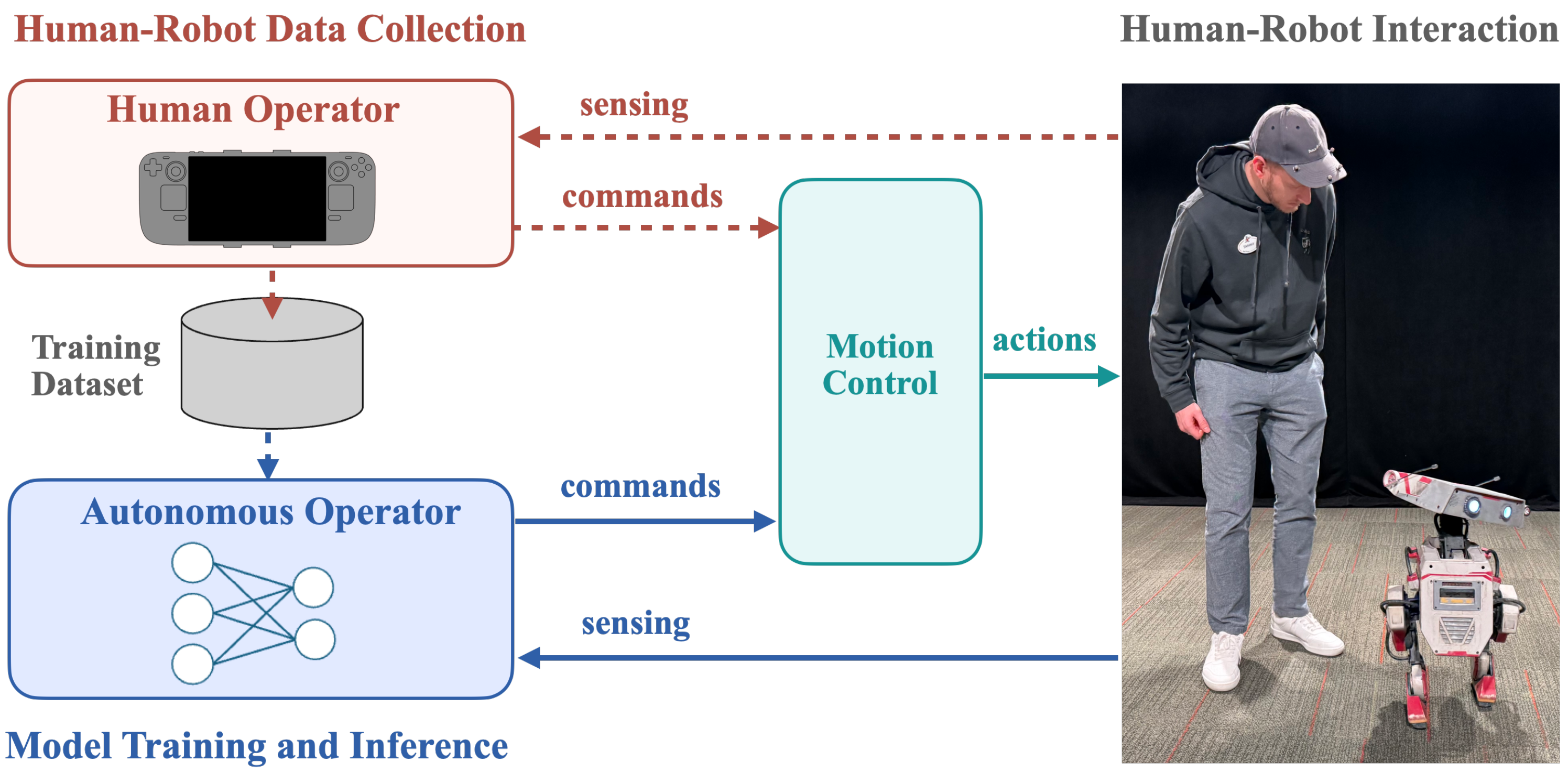} 
    \vspace{-5mm}
    \caption{\textbf{Overview of our approach.} We first collect a human-robot interaction dataset with a remote operator (red, top left). A motion control module takes operator commands and maps them to robot actions (cyan, middle). We then train a diffusion-based model that learns to imitate the operator (blue, bottom left). Our system learns to perform simple interactions with a human and express different moods. Dashed lines indicate components only required for data collection and training.}
    \label{fig:teaser} 
    \vspace{-6mm}
\end{figure}

In this work, we introduce a novel framework for HRI that learns to imitate operators. As shown in \figref{fig:teaser}, we first collect a human-robot interaction dataset, where an external operator controls the robot to interact with a human and express varying moods. The dataset includes the poses of the robot and the human, along with the operator's teleoperation commands. 
Crucially, the problem statement focuses on imitating the operator rather than the robot’s actions, offering several advantages over relearning low-level control. Leveraging existing motion control eliminates the need to relearn the complex and data-intensive process of the robot’s dynamics. Furthermore, it ensures the robot maintains its motion style, robustness, and safety constraints.

In our method, we use diffusion models to imitate operators and predict diverse interactions. We introduce several simple yet effective techniques to adapt these models for real robotic platforms and human-robot interactions. As opposed to previous motion diffusion frameworks, which primarily focus on continuous predictions \cite{chen2024camdm, tevet2023mdm}, we propose a model that can predict both continuous commands and discrete events, matching the layout of gamepads. Our method consists of a unified transformer backbone that integrates a diffusion-based module for continuous signal prediction and a classifier for discrete event prediction. To enable our model to react dynamically to human movement, we condition on the time-varying robot-relative human pose. Additionally, we propose masking input signals after encoding rather than before \cite{chen2024camdm}, since zero input signals can be a valid input, such as indicating the proximity of the human to the robot.

Our experiments show that with less than one hour of data, our framework learns to perform autonomous interactions, exhibit multiple moods, and switch between different control modes, such as walking and standing. In a user study, we let 20 users interact with our system and show that they struggle to distinguish between autonomous and operated behavior for simple interactions. Furthermore, we show that users can successfully recognize the different moods of the robot. Lastly, we demonstrate the zero-shot transfer of our models trained with interaction data collected with a non-anthropomorphic robot to a humanoid robot.  

In summary, our contributions are: 
\begin{itemize}
\vspace{-1mm}
    \item a novel system enabling simple autonomous human-robot interactions and expressing different moods using a small dataset of human operator data for training,
    \item a model to imitate operator commands based on human pose, predicting both continuous and discrete signals,
    \item and empirical evidence through simulated and real-world experiments that our model generates realistic human-robot interactions and transfers across robotic platforms with the same operator interface.
\end{itemize}

\section{Related Work}
We split related work into character control and learning-based HRI. Note that we consider our work as building up the machinery required for effective autonomous HRI, and there is extensive orthogonal research in social HRI that looks into the effects of robot personalities \cite{lee2006personality} and emotions \cite{paiva201421}. This opens the door for interesting future studies. We refer to \cite{sheridan2016human, robert2020review, kirby2010affectiverobots} for a detailed overview of social HRI.

\subsection{Character Control}
Learning-based controllers have shown remarkable success in controlling simulated and real-world characters or robots. Broadly, these controllers can be categorized into two types: \textit{Imitation-based} controllers, which rely on dense kinematic reference motions as input~\cite{fussell2021supertrack, luo2023phc, serifi2024vmp, liu2024mimicking, braun2023physically}, and \textit{goal-driven} controllers~\cite{grandia2024design, chen2024camdm, bergamn2019drecon, he2025hover}, which take high-level commands---such as joystick inputs---to control the character. These two paradigms are often closely intertwined. Many controllers leverage dense motion references during training, but exploit control through sparse user input at inference time~\cite{tessler2024maskedmimic}.

With the advances of diffusion models~\cite{ho2020denoising} and their applications in motion generation~\cite{tevet2023mdm}, recent work investigated their application in control settings. CAMDM~\cite{chen2024camdm} generates kinematic motion based on control signals in an autoregressive fashion. Closer to our work is the more direct intersection between text-conditioned diffusion models and physics-based controllers. RobotMDM~\cite{serifi2024robotmdm} generates motions tailored for a pretrained controller, whereas CLoSD~\cite{tevet2025closd} integrates a controller into the generation process. Our model builds on such physics-based approaches, but instead of dense references, we learn to imitate user inputs provided during data collection for human-robot interaction, which reduces the data requirement.

Diffusion models have also been explored in the context of HRI. Yoneda \etal~\cite{yoneda2023diffusha} propose a system for shared autonomy, where an autonomous agent assists a user in teleoperating a robot. Through the use of a diffusion model, the authors learn a model that trades off user autonomy for optimal behavior.
Diffusion Co-Policy~\cite{ng2023diffusion} is a method for a collaborative table moving task. Like them, we base our modeling on a diffusion model. However, in contrast to their approach, we support an action space with more than 2D continuous dimensions, supporting a combination of discrete \emph{and} continuous actions.  Moreover, our model controls the robot fully autonomously instead of controlling a subsystem that is attached to a robot commanded by a human.

\subsection{Learning-Based Human-Robot Interactions}
The application of deep learning techniques to autonomous HRI has recently gained popularity. Often, task-specific and isolated behaviors are investigated, such as human-robot handshakes~\cite{prasad2022human, christen2019drlhs}, human-robot handovers~\cite{christen2023handover, christen2024synh2r}, or table carrying~\cite{ng2023hrcc}. For a broader overview that includes non-learning-based techniques, we refer to~\cite{thomaz2016hri, sheridan2016human}.  

A large body of work has focused on using learning-based methods for physical human-robot interaction. A common approach is to leverage human-human interaction demonstrations. Nikolaidis \etal~\cite{nikolaidis2015efficient} propose an unsupervised learning algorithm to cluster human behavior and learn robot policies that align with the user. In~\cite{butepage2020imitating}, the authors introduce a deep generative model representing the joint distribution of interactions in latent space via variational auto-encoders. Similarly, some works focus on learning joint latent representations between the human and robot from demonstrations, such as for human-robot hand interactions~\cite{prasad2022mild} or social motion forecasting~\cite{valls2024echo}. In Co-GAIL~\cite{wang22cogail}, a human policy evolves along with a robotic policy via adversarial imitation learning. To adhere to the diversity of human behavior, a latent representation that represents the human's intent is used, similar to~\cite{xie2021learning} for multi-agent interactions. While there are similarities to these methods, our setting has different assumptions. In particular, we do not assume access to human-human demonstrations, as remotely operated robotic systems can be substantially different from the human anatomy, rendering a mapping between human demonstration and robot challenging. Furthermore, these methods typically focus on collaborative tasks with physical interactions, whereas we focus on non-physical interactions, such as following humans, expressing moods, and initiating pre-defined behaviors.

\section{Background Diffusion Models}
\label{sec:bg_diffusion}

Diffusion models~\cite{ho2020denoising} have become popular in various domains such as the generation of images~\cite{stablediffusion}, videos~\cite{ho2022video} or motion sequences~\cite{tevet2023mdm}. They are a family of generative models particularly well suited for modeling complex data distributions. Diffusion models comprise a forward and a backward process. The forward process is a discretization of Langevin dynamics, where in each step, Gaussian noise $\epsilon_t$ is added to a data sample $\mathbf{x}_0$, leading to progressively noisier versions of that sample \( \mathbf{x}_t \):
\begin{equation}
    \mathbf{x}_t = \sqrt{1-\beta_t} \mathbf{x}_{t-1}+\sqrt{\beta_t}\epsilon_t,\; \epsilon_t \sim \mathcal{N}(0,\mathbf{I}).
\end{equation} 
Recursively applying this update yields
\begin{equation}
    q(\mathbf{x}_t | \mathbf{x}_0) = \mathcal{N}(\mathbf{x}_{t}; \sqrt{\bar{\alpha}_{t}} \mathbf{x}_0, (1 - \bar{\alpha}_{t}) \mathbf{I}),
    \label{eq:noising}
\end{equation} 
\begin{equation}
    \bar{\alpha}_t = \prod\limits_{i=1}^{t} (1 - \beta_i),
\end{equation}
where \( \beta_i \) is a variance scheduler for the noise, with $i$ or $t$ indicating the diffusion step. In the reverse process, the data is gradually denoised into a clean sample, which is modeled by the probability distribution $p_{\theta}(\mathbf{x}_{t-1} | \mathbf{x}_t)$. This denoising step is learned via a neural network parameterized by $\theta$. We follow MDM \cite{tevet2023mdm} and directly predict the clean output $\mathbf{\hat{x}}_0$. Additionally, in conditional diffusion models, the probability distribution is extended by conditioning signals $\mathbf{c}$, yielding $p_{\theta}(\mathbf{x}_{0} | \mathbf{x}_{t}, \mathbf{c})$. During inference, the model can generate diverse outputs from randomly sampled Gaussian noise and the conditioning signals. 
\begin{figure}[t]
    \centering
    \includegraphics[width=1.0\columnwidth]{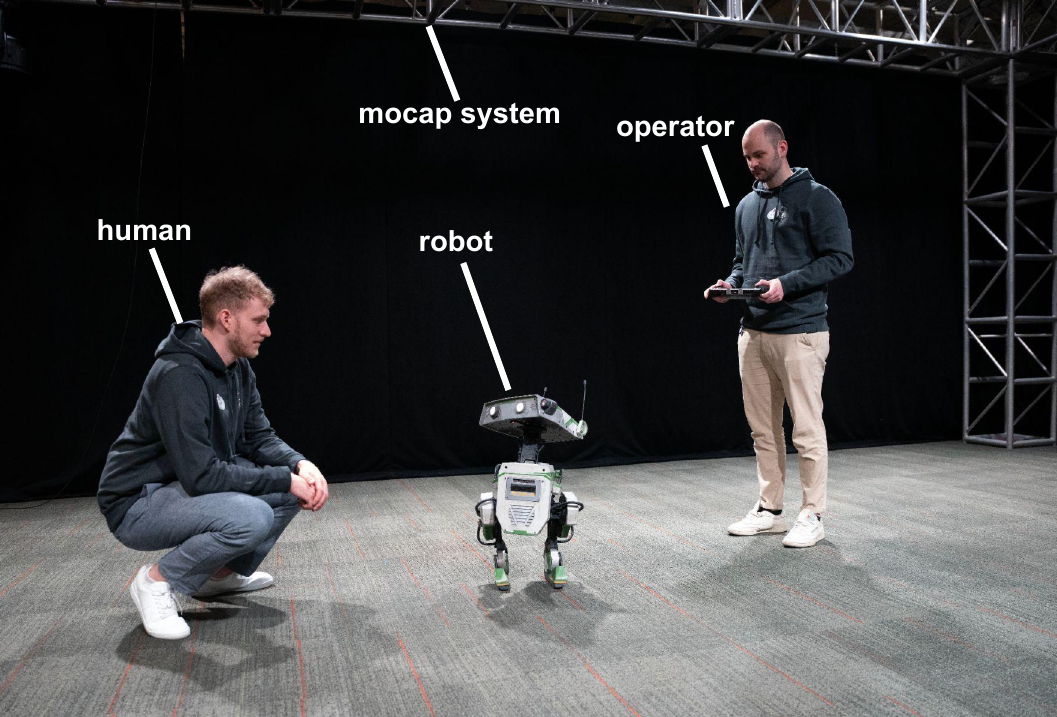}
    \caption{\textbf{Our capture setup.} An operator controls the robot to interact with a human participant. The poses of the human and robot, as well as the operator commands, are recorded in a motion capture studio.}
    \label{fig:capture} 
\end{figure}

\section{Problem Setting}
\label{sec:problem}
Given a sequence of past human poses $\mathbf{p} \in \R^{\text{M} \times \text{7}}$ and robot poses $\mathbf{r} \in \R^{\text{M} \times \text{7}}$, which each consist of M frames of global 3D positions and 4D orientations (quaternions), our goal is to predict a future sequence of continuous operator commands $\mathbf{x}^{\text{1:N}} \in \R^{\text{N} \times \text{j}}$, where N indicates the prediction horizon and j is the number of commands per frame. In addition to the continuous command predictions, such as joysticks on a gamepad, we are also interested in predicting discrete events, which are typically triggered through button presses on operator interfaces. We differentiate between discrete events that trigger a certain behavior $\mathbf{d}_{\textbf{b}}$, such as a pre-defined jumping or dancing motion (see \secref{sec:bg_bdx}), and the mode $\mathbf{d}_{\textbf{m}}$ of the robot, e.g., whether it should be in standing or walking mode. Each prediction window has one prediction per discrete input, rather than a separate prediction per frame. Both the continuous and discrete signals are passed as conditioning to a motion control module that controls the robot (see \secref{sec:bg_bdx}). Please note that we use subscripts to indicate the diffusion step and superscripts to describe the sequential prediction.

\section{Method}

Our framework consists of two main components: human-robot interaction data collection and a model to train autonomous interactions.
\subsection{Human-Robot Interaction Data Collection}
Our capture setup is shown in \figref{fig:capture} and consists of a robot, an operator controlling the robot, and a human interacting with it. We aim to capture the human poses $\mathbf{p}$, the robot poses $\mathbf{r}$, the continuous operator commands $\mathbf{x}$ as well as the discrete commands $\mathbf{d}_{\textbf{b}}$ and $\mathbf{d}_{\textbf{m}}$. We record operator commands directly on the input interface. To capture human and robot poses, we leverage the OptiTrack motion capture system \cite{optitrack}, placing markers on the robot and a hat worn by the human. 

To collect data, we ask an expert operator to perform interactions that are responsive to human pose, such as following the human and varying between expressing different moods through motions. For instance, in the shy mood, the robot tends to look at the ground while interacting with the human, whereas in the angry mood, the robot frequently shakes its head and refuses interactions when the human approaches. See \tabref{tab:dataset} for an overview of the collected data and a more detailed description of the moods. In our data collection, the robot is operated by a single expert operator and two humans, one after another, interact with it.

\subsection{Method Architecture}
\begin{figure}[t]
    \centering
    \includegraphics[width=\columnwidth]{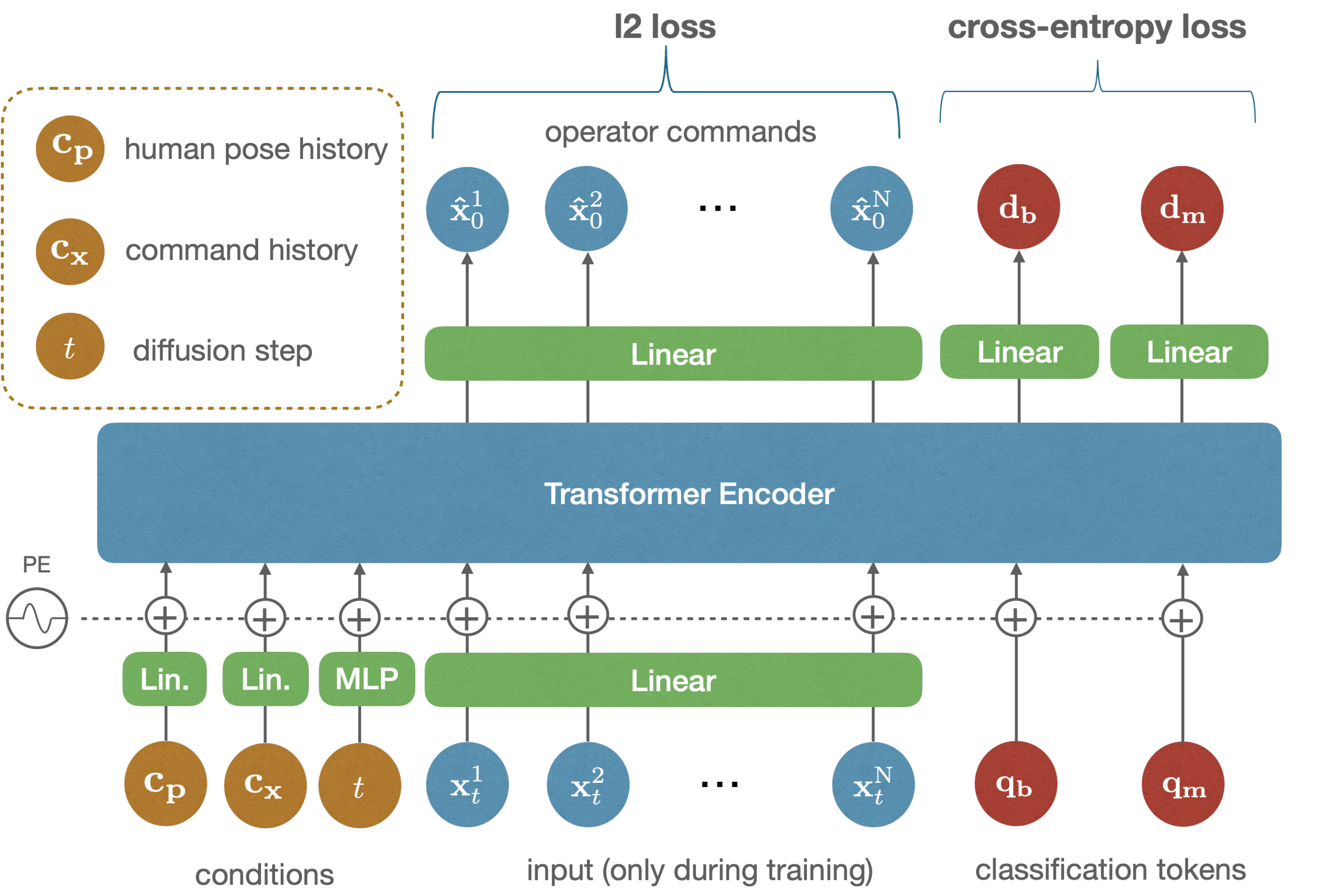} 
    \caption{\textbf{Overview of our method architecture.} We use conditions in the form of past human poses and commands (yellow). A diffusion model predicts operator commands to control the robot model (blue). The transformer also outputs discrete predictions for different behavior and the mode (red).}
    \label{fig:method} 
\end{figure}

Our method architecture is illustrated in \figref{fig:method}. Inspired by the CAMDM framework for character control in simulation \cite{chen2024camdm}, we propose a
 framework for real-world human-robot interaction. To address the challenges of human-robot interactions, we contribute several domain-specific adjustments and extensions to the architecture. 
Our model has a transformer backbone and is conditioned on multiple components; the history of human poses $\mathbf{c}_{\textbf{p}}$ and operator commands $\mathbf{c}_{\textbf{x}}$, and the diffusion step $t$. Before passing the history of human poses to the model, we transform them into a robot-relative frame using the history of robot poses $\mathbf{r}$.

In summary, our model $\mathcal{G}$ is defined as follows:

\begin{equation}
    (\mathbf{\hat{x}}_0, \mathbf{d}_{\textbf{b}},\mathbf{d}_{\textbf{m}}) = \mathcal{G}(\mathbf{c}_{\textbf{p}}, \mathbf{c}_{\textbf{x}}, t, \mathbf{x}_t, \mathbf{q}_{\textbf{b}}, \mathbf{q}_{\textbf{m}}).
\end{equation}

\subsubsection{Continuous Command Prediction via Diffusion}
To predict continuous operator commands, we use a diffusion model (see \secref{sec:bg_diffusion}). More specifically, during training, we alternate between noising clean data samples $\mathbf{x}_0$ to noisy signals $\mathbf{x}_t$ according to Eq. \ref{eq:noising} and denoising steps to predict the clean signals $\mathbf{\hat{x}}_0$. 

We apply the separation of conditioning tokens and classifier-free guidance on past commands as proposed by \cite{chen2024camdm}. In contrast to their model, we found that dropout in the positional encoding leads to noisy outputs, hence do not adopt it in our model. Furthermore, the conditioning signals in our case can be close to zero, for example, when the human is close to the robot. Therefore, we propose to apply masking to the conditioning \emph{after} the encoding layers instead of to the raw conditions \cite{chen2024camdm}, because otherwise zero-masking could be mistaken by the model for actual zero conditions. To account for the diversity in human height, we augment the available human pose data by adding a random offset ($\pm$ \SI{0.3}{\meter}) in the negative gravity direction during training. 

\subsubsection{Discrete Event Prediction}
It is non-trivial to mix continuous and discrete signals in a diffusion process. We propose to add the discrete event predictions as auxiliary tasks to our transformer model. In particular, we add classification query tokens, specifically $\mathbf{q}_{\textbf{b}}$ for discrete behavior and $\mathbf{q}_{\textbf{m}}$ for mode, to the transformer architecture \cite{dosovitskiy2020image}. The output of the transformer encoder has a head to predict the mode $\mathbf{d}_{\textbf{m}}$  and pre-defined discrete behavior $\mathbf{d}_{\textbf{b}}$, respectively. The model learns to select from several classes of behaviors and a default class (i.e., no discrete event occurring). Note that these predictions use the same transformer encoder, but no diffusion is applied. To train the classification heads, we use weighted cross-entropy losses, due to the imbalance in the data between the rare discrete events and the default class.

\begin{table}[t!]
\centering
\caption{Motion Types, Lengths, and Descriptions}
\label{tab:dataset}
\begin{tabularx}{\columnwidth}{l c X}
\toprule
\textbf{Category} & \textbf{Length (min)} & \textbf{Description} \\
\midrule
Default  & 8 & Follows the human, retreats if the human walks toward it, looks at the human when in standing mode. \\
Angry    & 6 & Ignores the human, walks away if approached, shakes its head, occasionally triggers angry animations. \\
Sad      & 8 & Walks away from the human, shakes its head, looks mostly down at the ground. \\
Shy      & 7 & Approaches the human carefully, occasionally stops, mostly looks at the ground, and tries to avoid eye contact. \\
Happy   & 8 & Turns in circles, runs after the human, and often expresses positivity through animations such as dancing or jumping. \\
\bottomrule
\end{tabularx}
\end{table}

\begin{figure*}[t!]
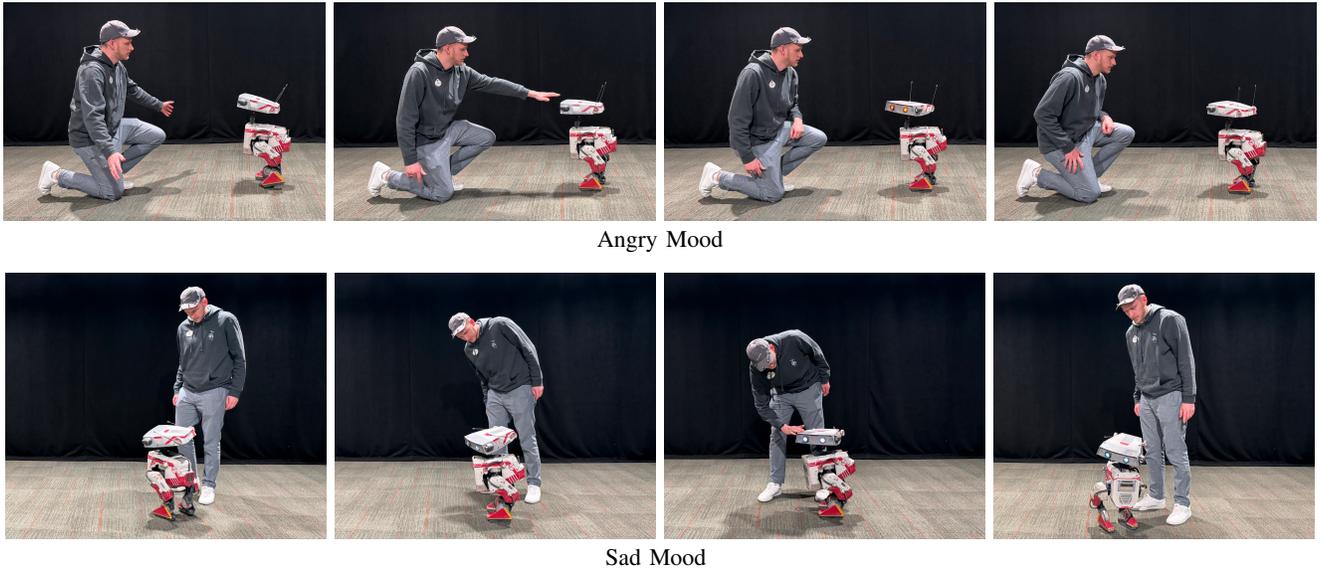

 \centering
  \begin{minipage}{\linewidth}
  \centering
  \begin{minipage}{0.24\linewidth}
   \centering
   \includegraphics[width=\linewidth,trim={600 400 1700 500},clip]{sections/figures/angry/angry_us_1.pdf}
   \\ 
  \end{minipage}
  \begin{minipage}{0.24\linewidth}
   \centering
   \includegraphics[width=\linewidth,trim={600 400 1700 500},clip]{sections/figures/angry/angry_us_2.pdf}
   \\ 
  \end{minipage}
  \begin{minipage}{0.24\linewidth}
   \centering
   \includegraphics[width=\linewidth,trim={600 400 1700 500},clip]{sections/figures/angry/angry_us_3.pdf}
   \\ 
  \end{minipage}
  \begin{minipage}{0.24\linewidth}
   \centering
   \includegraphics[width=\linewidth,trim={600 400 1700 500},clip]{sections/figures/angry/angry_us_4.pdf}
   \\ 
  \end{minipage}
 \end{minipage}
 \\~\\ \vspace{0mm}
   \small{Angry Mood}
    \\~\\ \vspace{0mm}
  \begin{minipage}{\linewidth}
  \centering
  \begin{minipage}{0.24\linewidth}
   \centering
   \includegraphics[width=\linewidth,trim={2000 200 200 100},clip]{sections/figures/sad/sad_us_1.pdf}
   \\ 
  \end{minipage}
  \begin{minipage}{0.24\linewidth}
   \centering
   \includegraphics[width=\linewidth,trim={2000 200 200 100},clip]{sections/figures/sad/sad_us_2.pdf}
   \\ 
  \end{minipage}
  \begin{minipage}{0.24\linewidth}
   \centering
   \includegraphics[width=\linewidth,trim={2000 200 200 100},clip]{sections/figures/sad/sad_us_3.pdf}
   \\ 
  \end{minipage}
  \begin{minipage}{0.24\linewidth}
   \centering
   \includegraphics[width=\linewidth,trim={2000 200 200 100},clip]{sections/figures/sad/sad_us_4.pdf}
   \\ 
  \end{minipage}
  \end{minipage}
  \\~\\ \vspace{0mm}
   \small{Sad Mood}
\vspace{-1mm}
 \caption{\textbf{Example behavior of different moods.} In the angry mood, the robot refuses interactions and steps away from the human (top). In the sad mood, the robot turns away from the human, has its head tilted towards the ground and occasionally shakes its head (bottom).}
 \label{fig:qual}
\end{figure*}

\section{Experiments}
In \secref{sec:bg_bdx}, we describe the robotic platform used in our experiments and provide implementation details about model parameters and runtime of our approach in \secref{sec:impl}. We conduct experiments in simulation in \secref{sec:sim}. In \secref{sec:user}, we perform a user study in the real world and demonstrate zero-shot transfer to a different robot in \secref{sec:zero}. See our video\footnote{\url{https://youtu.be/4U4etupwzhQ?si=NNHM7NqjAKWoo32B}} for more qualitative results.

\subsection{Robotic Platform}
\label{sec:bg_bdx}

We build on the platform developed by Grandia \etal~\cite{grandia2024design}, which proposes a new bipedal character design and control approach for entertainment applications. A reinforcement learning based control architecture is used to robustly imitate artistic motions conditioned on continuous and discrete command signals. During runtime, these command signals are generated by an animation engine that fuses user-inputs with predefined animation content. An intuitive operator interface, implemented on the hand-held operator controller, enables expressive real-time show performances with the robot. 

The robot, \SI{0.66}{\meter} tall, with a total mass of \SI{15.4}{\kilo\gram}, has 5 degrees of freedom (DoF) per leg and a 4 DoF neck. Moreover, the robot is equipped with a set of show functions: a pair of actuated antennas and illuminated eyes, and a headlamp. Additionally, the robot has a stereo pair of loudspeakers in both the body and the head. These components provide additional means to express moods. Their behavior is synchronized with the motion of the robot through animation signals from the animation engine and state feedback. We refer the reader to \cite{grandia2024design} for more details.

\subsection{Implementation Details}
\label{sec:impl}
Our transformer encoder has a latent dimension of 128, a feedforward size of 256, and two attention heads across 2-layers. We train for 5000 epochs with a batch size of 128 and a learning rate of 1e-4. We select a past window of 15 frames and predict 25 future frames (M and N in \secref{sec:problem}). The number of diffusion steps is set to 8. We predict 10 continuous signals, up to 8 discrete behaviors (depending on the mood), and 2 modes (walking and standing). To improve signal quality, we apply a simple Gaussian smoothing filter to the model's predictions before sending them to the robot. This helps to reduce the noise introduced by the motion capture conditioning.

All computations, including our diffusion model, motion control policy, state estimation, and the animation engine, run on the robot's on-board computer. Both human pose and autonomous operator commands are sent to our model at \SI{50}{\hertz}. The predicted operator commands are passed to the animation engine (see 
\secref{sec:bg_bdx}), which fuses the commands with animation content and sends it to a control policy. This low-level control policy outputs actuator setpoint commands, which are interpolated to \SI{600}{\hertz} and sent to the actuators' PD controllers. In addition to teleoperation commands, the low-level policy also receives proprioceptive state inputs from the robot.

\subsection{Simulation-Based Evaluation}
\label{sec:sim}
Quantifying engaging and natural interactions in simulation is challenging, much like defining general heuristics for robot control or rewards for learning behavior. Nevertheless, to measure and compare architectural choices, we define a set of metrics that provide insights into model performance:

\textbf{Facing Angle Error (FAE):} measures the average angle between the robot's forward direction and the vector from the robot's root to the human in degrees. We assume that the robot should be facing the human (in the default mode).

\textbf{Tracking Error (TE):} measures the average distance between the human and the robot in the x-y plane. We assume that the robot should follow the human closely.

\textbf{Mean Squared Derivative (MSD):} measures how rapidly the continuous signals change. This helps identify noisy predictions and abrupt transitions between prediction windows. We average over all signals. \\

\begin{figure}[t]
    \centering
    \includegraphics[width=0.6\linewidth]{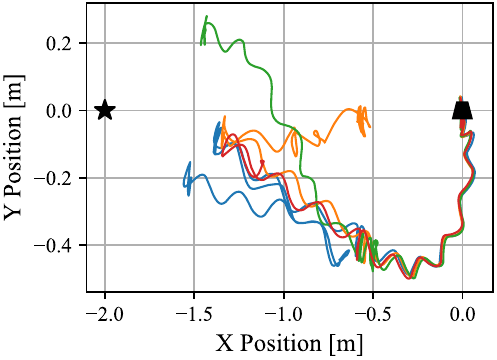} 
    \vspace{-2mm}
    \caption{\textbf{Diversity of our framework.} Given the same starting point (black trapezoid) and a fixed human position (black star), we run our model multiple times and plot x-y positions. As can be seen, the model generates different behavior for the same human pose conditions.}
    \label{fig:qual_diversity} 
\end{figure}

We compare our approach to a transformer baseline and ablate design choices of our architecture. For our simulation experiments, we use a 75/25 train-test split for the default mode. During evaluation, we replay human motions from the test set and control the robot using our model, streaming the human-relative pose from simulation directly to our model. Additionally, as shown in \figref{fig:qual_diversity}, we qualitatively demonstrate that, given a fixed human pose and a single initial robot pose, our model generates diverse outputs. 
\subsubsection{Baseline Comparison} 
Our baseline is a deterministic model that directly maps from human pose as input to operator commands as output. To achieve this, we use the same transformer architecture that is used in our model but remove the diffusion process. To account for the sequential output, we add query tokens as inputs. As can be seen in \tabref{tab:ablations} (top), our model achieves better tracking of the human (lower FAE and TE) and leads to a cleaner signal and transitions between prediction windows, as indicated by the lower MSD score. This shows that using a diffusion model is beneficial over using a transformer model without diffusion. 
\subsubsection{Ablations}
To ablate the components of our architecture, we train different variants of our model by removing the human pose history $\mathbf{c}_{\textbf{p}}$ (\textit{w/o human}), command history $\mathbf{c}_{\textbf{x}}$ (\textit{w/o commands}), dropout (\textit{w/o dropout}). Additionally, we evaluate the influence of different prediction window lengths (25, 50, and 75 frames). We present results in \tabref{tab:ablations}. As expected, the variant without information about the human pose leads to high tracking errors. The variant without conditioning on past commands leads to the best tracking of the human, because no constraints are imposed on the coherence of the signal to previous predictions. Hence, the transitions between windows are often abrupt and jerky, as indicated by the high MSD (14.8). Our final variant (Ours 25 frames) trades off signal coherence with close tracking of the human. When comparing different prediction windows, we find that 25 frames yield the best tracking performance while staying close to real-time, considering the computational load of the diffusion process. Note that all but the window size ablations are trained with a window of 25 frames.

\begin{table}[t!]
    \centering
     \caption{Simulation-based Evaluations}
    \begin{tabular}{lccc}
        \toprule
        Variant & FAE [\SI{}{deg}.] $\downarrow$ & TE [\SI{}{m}] $\downarrow$ & MSD $\downarrow$ \\
        \midrule
        \textit{transformer} & ~57.66 $\pm$ 19.61 & 1.49 $\pm$ 0.05  & 4.40 $\pm$ 2.29 \\
        \textit{Ours 25 frames} & ~43.85 $\pm$ ~2.40 & 1.47 $\pm$ 0.03 & 2.42 $\pm$ 0.40 \\ \hline
        \textit{w/ dropout}  & ~39.48 $\pm$ ~2.19 & 1.44 $\pm$ 0.02 & 5.76 $\pm$ 0.32 \\
        \textit{w/o human} & 100.28 $\pm$ ~4.74 & 3.20 $\pm$ 0.50 & 2.34 $\pm$ 0.26 \\
        \textit{w/o commands} & ~43.12 $\pm$ ~2.66 & 1.44 $\pm$ 0.01 & 14.80 $\pm$ 0.69\\
        \textit{Ours 75 frames} & ~56.97 $\pm$ ~1.98 &  1.54 $\pm$ 0.01 & 4.11 $\pm$ 0.39 \\
        \textit{Ours 50 frames} & ~50.58 $\pm$ 12.31 & 1.48 $\pm$ 0.02 & 2.45 $\pm$ 0.33 \\
        \textit{Ours 25 frames} & ~43.85 $\pm$ ~2.40 & 1.47 $\pm$ 0.03 & 2.42 $\pm$ 0.40 \\
        \bottomrule
    \end{tabular}
    \label{tab:ablations}
    \vspace{-2mm}
\end{table}

\subsection{Real-World Evaluation}
\label{sec:user}
We perform qualitative and quantitative real-world evaluations. Please see Fig.~\ref{fig:qual} and our accompanying video for qualitative results. In our empirical evaluations, we aim to answer whether our model can enable 1) interactions with the robot that feel similar to interactions controlled by a trained operator and 2) a robot to express different moods that are recognizable by humans. To analyze these questions, we propose a two-stage in-person user study. We run the user study with 20 participants, aged between 22 and 44 (M=30, SD=4.8), 12 males and 8 females. The participants were recruited from our organization, but were neither part of the project nor had experience interacting with our robot. After the interactive part of the study, we let participants fill out a small qualitative questionnaire.

\subsubsection{Questionnaire}
The anonymous questionnaire comprises three statements. The users are asked how much they agree with each statement on a 5-point scale, with a score of 1 indicating ``strongly disagree'' and 5 indicating ``strongly agree''. The first statement is ``I have experience interacting with robots'', which received a mean score of 2.2 (STD=1.0), showing that participants are not very experienced interacting with robots. The other statements, ``The interaction felt engaging and enjoyable'' and ``I felt like the robot was reacting to me'' received mean scores of 4.9 (STD=0.3) and 4.7 (STD=0.5), respectively. These scores indicate that participants enjoyed interacting with the robot and felt like the robot was reacting to their movements.  

\subsubsection{Operator Recognition}
In this experiment, we follow a protocol where each participant experiences one of two settings. In the first setting, the operator is controlling the robot to interact with the human in the default mode, following the human without expressing different moods. In the second setting, our model controls the robot to exhibit the same behavior. We demonstrate each setting to the user for 30 seconds, after which the participant has to guess whether it was our model or the operator controlling the robot. In total, we show each setting twice (4 trials in total) in randomized order.
To avoid bias, the operator pretends to be actively controlling the robot, even in the autonomous setting. We report the recognition accuracy in a confusion matrix shown in \tabref{tab:conf_matrix_operator}, where rows indicate the presented setting (GT) and columns are the user-predicted setting (User). As can be seen, the accuracy is close to 50\%, with scores of 55\% for the active control by the operator and 54\% for our autonomy mode. The autonomous model is classified as operator 46\% of the time and the operator is classified as autonomous model 45\% of the time. These results indicate that it is difficult for users to tell the two models apart, and that our model in default mode is close to an expert operator's abilities. Three participants, who were more experienced with robots, figured out tricks to distinguish between the models. For example, these participants realized that if we lose track of the human at the border of the mocap space, the model will only start following once inside the tracked space.

\begin{table}[t]
\centering
\caption{Operator Recognition
Confusion Matrix}
\begin{tabular}{lcc}
\toprule
GT \textbackslash~User & Operator & Autonomous \\
\midrule
Operator     & \colorcell{0.55} & \colorcell{0.45} \\
Autonomous   & \colorcell{0.46} & \colorcell{0.54} \\
\bottomrule
\end{tabular}

\label{tab:conf_matrix_operator}
\end{table}

\begin{table}[t]
\centering
\caption{Mood Recognition Confusion Matrix}
\begin{tabular}{lcccc}
\toprule
        GT \textbackslash User                             & Happy & Sad & Angry & Shy \\
\midrule
Happy & \colorcell{0.74} & \colorcell{0.00} & \colorcell{0.11} & \colorcell{0.16} \\
Sad   & \colorcell{0.00} & \colorcell{0.74} & \colorcell{0.05} & \colorcell{0.21} \\
Angry & \colorcell{0.26} & \colorcell{0.00} & \colorcell{0.74} & \colorcell{0.00} \\
Shy   & \colorcell{0.05} & \colorcell{0.11} & \colorcell{0.16} & \colorcell{0.68} \\
\bottomrule
\end{tabular}

\label{tab:conf_matrix}
\end{table}

\begin{figure*}[t!]
   \centering
  \begin{minipage}{\linewidth}
  \centering
  \begin{minipage}{0.245\linewidth}
   \centering
   \includegraphics[width=\linewidth,trim={300 0 300 100},clip]{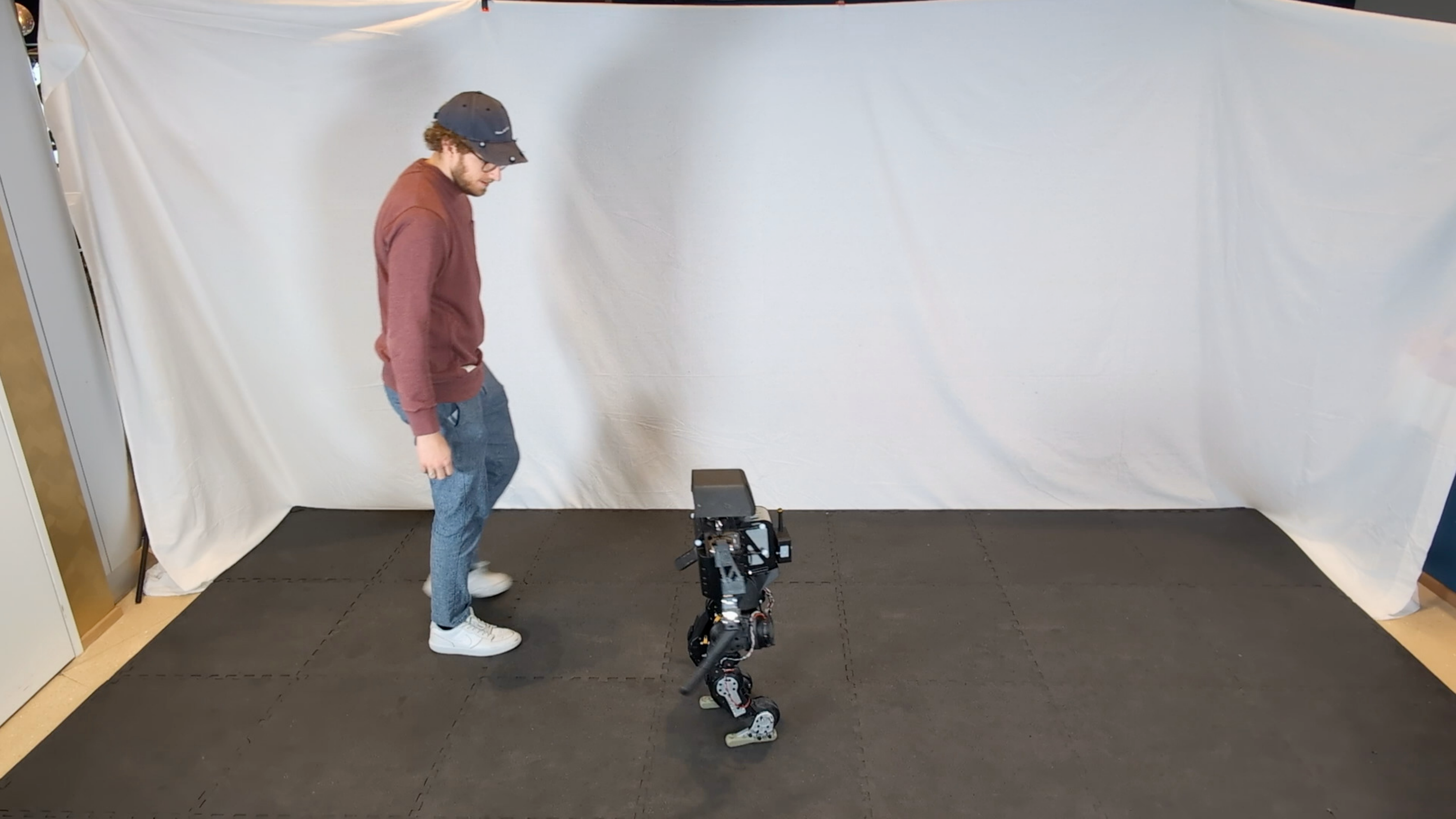}
   \\ 
  \end{minipage}
  \begin{minipage}{0.245\linewidth}
   \centering
   \includegraphics[width=\linewidth,trim={300 0 300 100},clip]{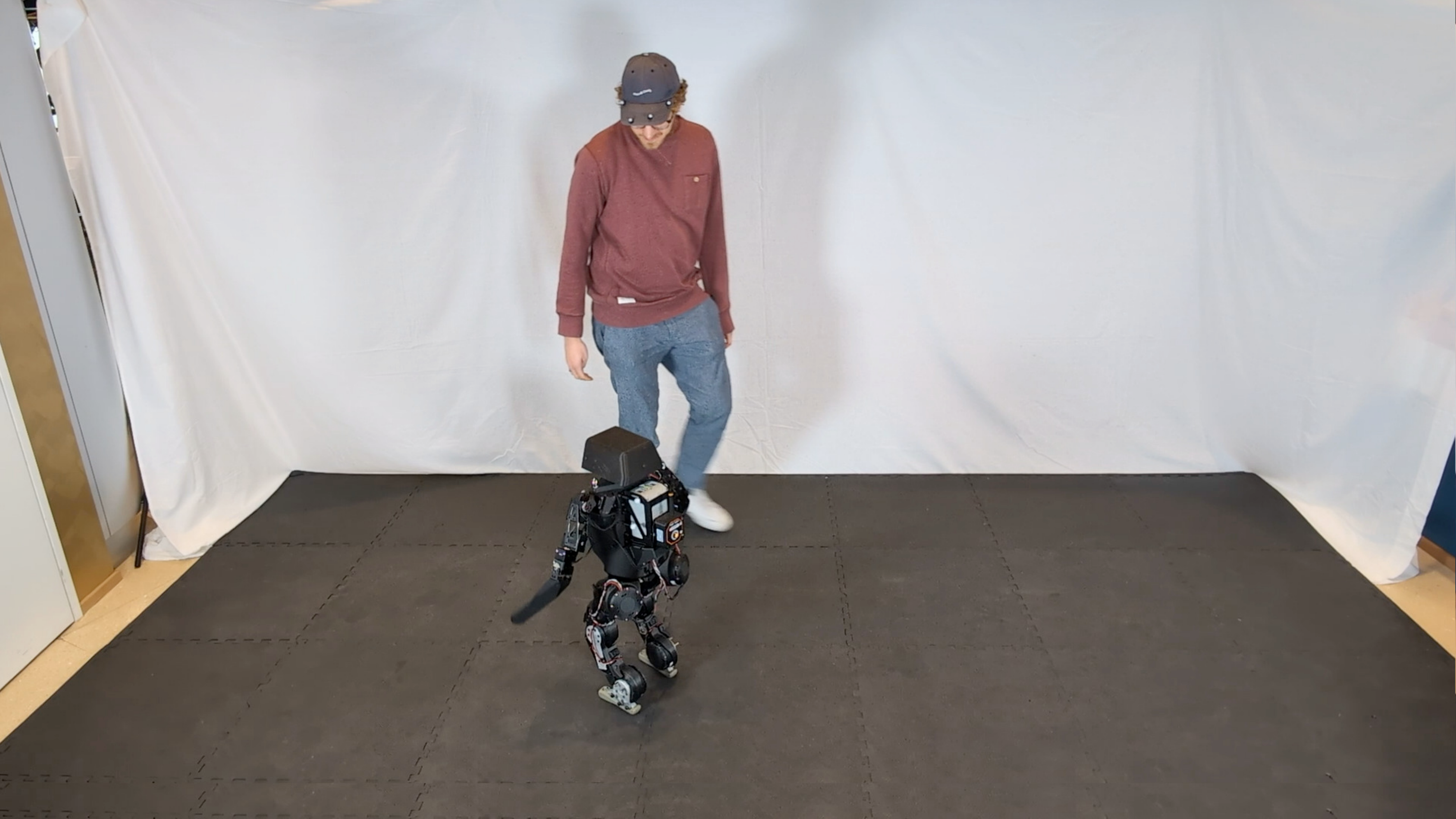}
   \\ 
  \end{minipage}
  \begin{minipage}{0.245\linewidth}
   \centering
   \includegraphics[width=\linewidth,trim={300 0 300 100},clip]{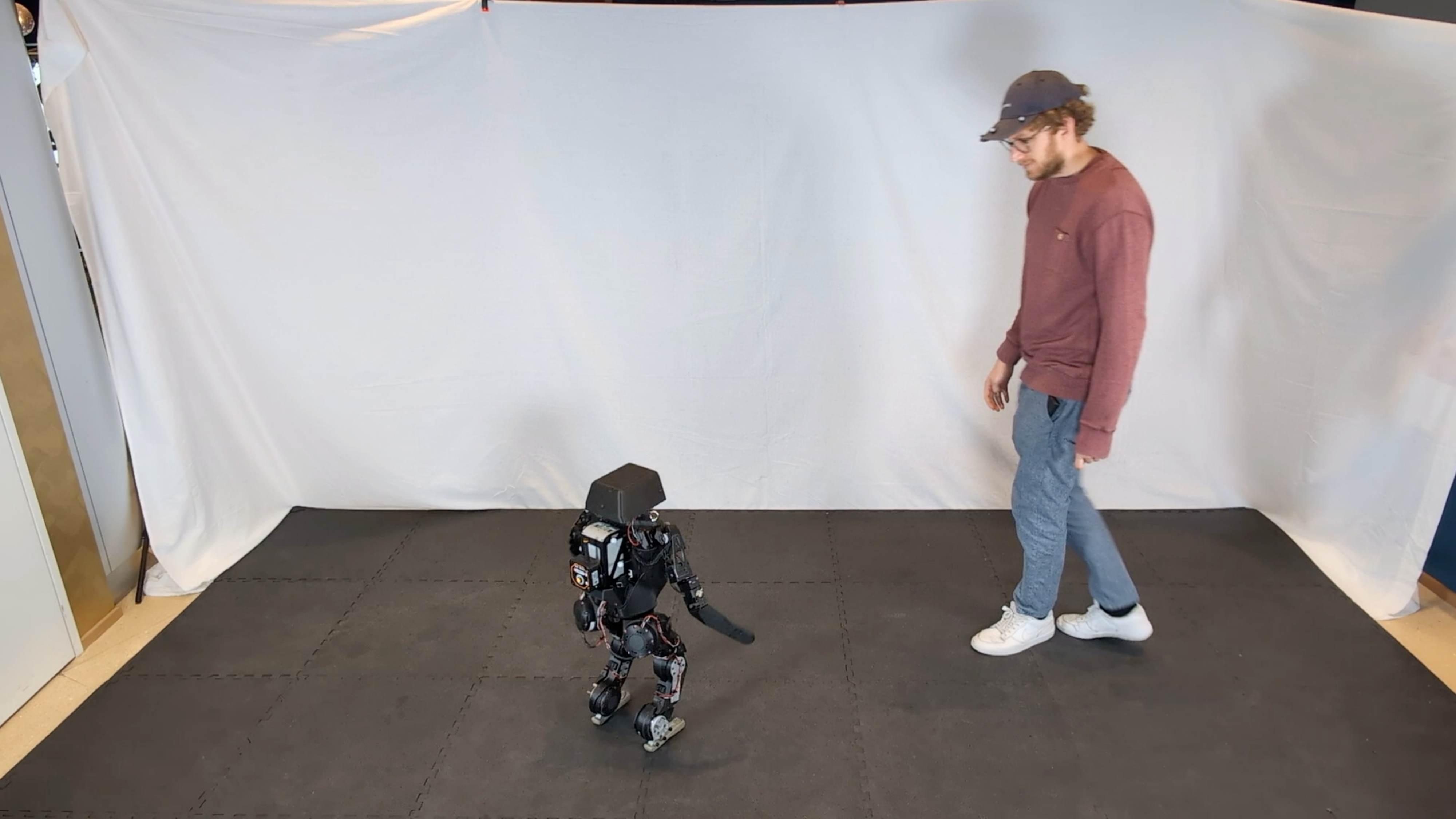}
   \\ 
  \end{minipage}
  \begin{minipage}{0.245\linewidth}
   \centering
   \includegraphics[width=\linewidth,trim={450 0 150 100},clip]{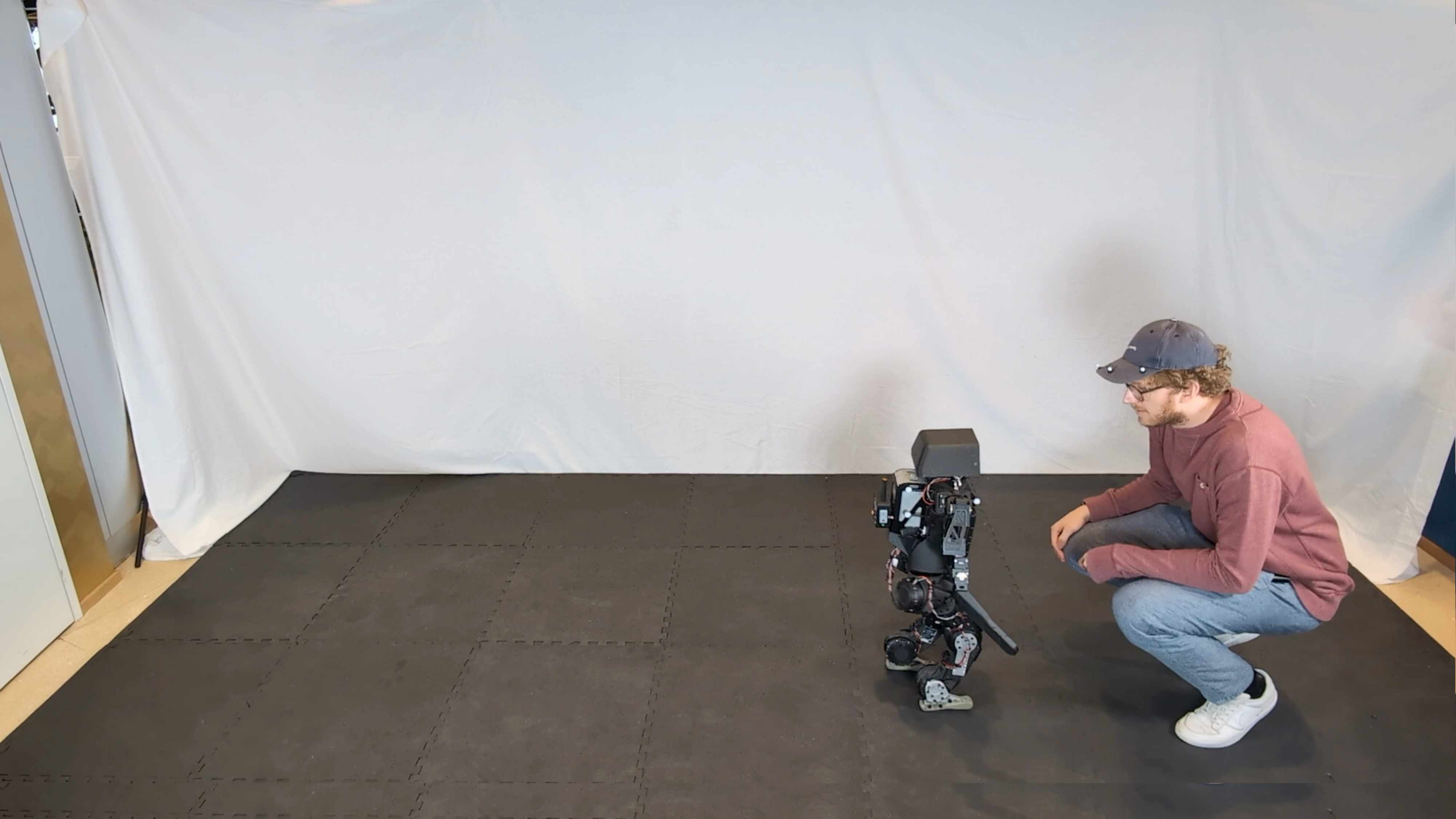}
   \\ 
  \end{minipage}
  \end{minipage}

 \caption{\textbf{Zero-shot transfer to a humanoid robot.} Our method trained with data collected via our non-anthropomorphic bipedal robot can successfully transfer behavior such as tracking the human without retraining.}
 \label{fig:lima}
\end{figure*}

\subsubsection{Mood Recognition} 
To determine whether the robot can express moods through our approach, we demonstrate the four moods (happy, sad, angry, shy) to the humans (see \tabref{tab:dataset} for a detailed description). We randomize the order of moods between participants and let them interact with the robot. An interaction takes one minute, after which humans have to provide a forced-choice answer. We do not explain the moods in detail and simply provide the moods to the humans. To avoid external bias, we use neutral eye coloring during this experiment (as opposed to the colored eyes for different moods, cf. \figref{fig:qual}). We report results in a confusion matrix shown in \tabref{tab:conf_matrix}, where rows indicate the presented setting (GT) and columns are the user-predicted setting. Generally, the users were able to correctly select the presented mood with an accuracy between 68\% and 74\%. The confusion cases give some interesting insights into the perception of certain moods. The angry mood was considered happy (26\%) because certain discrete behaviors, such as body shaking in disagreement, were considered a shake that expresses excitement (cf. video \SI{35}{\second}). In contrast, the happy mood was considered angry by some (11\%), because of the sprint towards the human, which was considered as an aggressive ``charging'' motion (cf. video \SI{1}{\minute}\SI{45}{\second}). The sad mood was misclassified as shy (21\%) due to the robot mostly looking at the ground. We believe these results validate that users can reliably identify the moods, with reasonable explanations for the confused cases. This also shows that there is no clear line between moods, and certain behaviors are interpreted differently between humans.

\subsection{Zero-Shot Cross-Embodiment Transfer}
\label{sec:zero}
To demonstrate that our method is applicable to different robotic platforms, we perform zero-shot transfer to a humanoid robot. Importantly, the interface mapping is maintained between the two platforms, i.e., the same joystick is used for controlling the walking direction and speed of the two robots. We qualitatively show in \figref{fig:lima} and our supplemental video that the default behavior, such as tracking the human, turning, and walking backwards (cf. \tabref{tab:dataset}), can be successfully transferred to a different platform. Notably, the training data used for our model was collected using the robot presented in \secref{sec:bg_bdx}. 

\section{Discussion}
While our approach takes a step towards autonomous human-robot interactions, it has several limitations. The main limitation is the reliance on a motion capture system to sense the robot-relative human pose. In the future, we hope to integrate perception to enable deployment in real-world environments. Beyond estimating global human pose, future work could predict the full-body human pose or even facial expressions, enabling more nuanced interactions. For example, we manually trigger the different moods in this work. A promising direction is exploring a single model conditioned on detailed human states to autonomously predict mood transitions. Furthermore, we do not model complex behaviors with long-term dependencies and high-level decision-making, such as combining multiple semantically different interactions into a cohesive interaction. Instead, our focus is on achieving performance comparable to human operators for shorter and simpler interactions, such as expressing moods and reacting to human pose. For more complex behavior, future research could explore the integration of a high-level decision-making module that links multiple interactions together.

Our data was captured with a single operator and two human subjects. Our user study revealed a large diversity in how humans interact with the robot. Expanding the dataset with more participants could improve the model's robustness to diverse human behavior while increasing the diversity of interactions our model predicts. This would also enhance the adaptability of our model to different human heights, which we currently address through height randomization during training. Future work could also explore interactions that involve physical contact. Finally, current interactions are limited to one robot and one human. An exciting avenue for future research is extending our work to multi-robot and multi-human interactions.

\section{Conclusion}

We have proposed a method that enables autonomous human-robot interactions by learning the mapping from human pose to operator commands with a diffusion-based approach, predicting both continuous commands as well as discrete button presses. By relying on operator instead of actuator commands, we train a model with less than 40 minutes of data, drastically reducing data requirements. Our approach also enhances safety, as operator interfaces are typically designed with built-in safety guarantees. Lastly, our experiments demonstrate that users enjoy interacting with the system, can reliably recognize different moods, and find it difficult to distinguish between autonomous or operator-controlled robot behavior for simple interaction.

\section{Acknowledgements}
We thank Jenny Wang for her exploratory work. We would also like to thank all participants that took part in our user study and provided feedback on our system.

\bibliographystyle{IEEEtran}

\bibliography{IEEEabrv,references}

\end{document}